\documentclass{llncs}
\usepackage{llncsdoc}  

\usepackage{graphics} 
\usepackage{epsfig} 
\usepackage{mathptmx} 
\usepackage{times} 
\usepackage{amsmath} 
\usepackage{amssymb}  
\usepackage{subfigure}

\begin{document}
\title{
Collision Avoidance of Two Autonomous Quadcopters 
}
\author{Michalis Smyrnakis\inst{1} \and Jonathan M. Aitken\inst{1} \and Sandor M. Veres\inst{1} \thanks{This work has been supported by the Engineering and Physical
Sciences Research Council under grants EP/J011894/2 and EP/J011770/1}}

\institute{Department of Automatic Control and Systems Engineering, University of Sheffield \\ \email{m.smyrnakis, jonathan.aitken, s.veres@sheffield.ac.uk}}

\maketitle
\thispagestyle{empty}
\pagestyle{empty}

\begin{abstract}
Traffic collision avoidance systems (TCAS) are used in order to avoid incidences of mid-air collisions between aircraft. We present a game-theoretic approach of a TCAS designed for autonomous unmanned aerial vehicles (UAVs). A variant of the canonical example of game-theoretic learning, fictitious play, is used as a coordination mechanism between the UAVs, that should choose between the alternative altitudes to fly and avoid collision. We present the implementation results of the proposed coordination mechanism in two quad-copters flying in opposite directions. 
\end{abstract}


\section{Introduction}
\label{sec:intro}

In this paper we present a Traffic Collision and Avoidance System (TCAS), motivated by interaction between two quadcopters. Crucially this novel approach does not require any communication between the agents controlling the copters, cooperation is generated by a game-theoretic approach using physical observations and predictions of the action of the other vehicle, obtained solely by visual monitoring of the environment.

Unmanned aerial vehicles (UAVs) present a unique challenge in robots. A ground robot can hold position indefinitely, simply by stopping. For flying vehicles this is a very different preposition. Even though a quadcopter can hover in the air this consumes battery power, so prolonged processing is not an efficient solution. As technology develops UAVs will be used for a wide range of different tasks including inspection work, for example hard to access structures~\cite{Inspection}. The focus will be on operating the quadcopters autonomously to minimise human workload.

Naturally, as part of this task, the quadcopters will need to avoid each other as they move around in the environment. Communication between each quadcopter provides a potential solution, but creates an exponentially increasing overhead as the number of quadcopters increases, and vastly increases power considerations on-board. By reducing communication overhead, we can conserve power to lengthen the time spent on the task. Therefore each should be able to take decisions itself based on what it observes - game theory provides the potential for this by allowing each agent to predict the actions of others without requiring communication.

TCAS~\cite{Williamson} is a system designed to prevent collision between aircraft. Should two aircraft be on a collision course, a warning is given to one pilot with a recommendation to climb or descend, and an equivalent warning with an opposite instruction given to the other pilot. This is an efficient system enabling collision avoidance, but requires communication of information between both parties, if contradicting information is given, for example in the case of the \"{U}berlingen mid air collision where one aircraft acted on TCAS and the other on instructions from a ground controller, then the probability of an accident is significantly increased~\cite{Aitken10}.

In this work we present a very different solution based on the game theoretic approach. Whilst similar techniques use off-line learning~\cite{Johnson} to establish behaviours, this approach takes decisions on-line, requiring no a priori training, no communication between the players and presents a system which generates agreement online; allowing the process of cooperation itself to be observed.


\section{Problem definition}
\label{sec:probdef}

In this section we define the TCAS problem as a symmetric game where the UAVs are the players of the game and the altitudes that they can fly are the actions available to each player. We start by briefly providing some basic game-theoretic definitions that we will use throughout this paper and then we show how the TCAS problem can be cast as a game. 

\subsection{Game-theoretic definitions}
The strategic game used in this paper has a set of players, $i=1,\ldots\mathcal{I}$, who choose their actions from their actions sets $S^{i}$ simultaneously. We will often write $-i$ in order to refer to all players but $i$.  Each player $i$ chooses his action, $s^{i}\in S^{i}$. We will refer to the actions that all the players chose in a game as joint action $s=(s^{1},\ldots,s^{\mathcal{I}}) \in S=\times_{i=1}^{\mathcal{I}}S^{i}$. The rewards that players gain is a mapping from the joint action space to the real numbers, $r^{i}(s): S\rightarrow \mathbb{R}$. 

The players of a game choose their actions using mixed strategies. A mixed strategy of a player $i$ is defined as $\sigma^{i} \in \Delta^{i}$, where $\Delta^{i}$ is the set of all probability distributions over the action space  $S^{i}$ of player $i$. Similarly to actions, a joint mixed strategy $\sigma=(\sigma^{1},\ldots \sigma^{\mathcal{I}}) \in \Delta=\times_{i=1}^{\mathcal{I}}\Delta^{i}$, is the set of all the mixed strategies players use in a game. Special cases of mixed strategies are pure strategies. A player that chooses an action with probability one chooses his actions using a pure strategy. 

Players can use any decision rule in order to choose their mixed strategy in the game. A common decision rule in game-theoretic literature is Best Response $(BR)$. When players are using $BR$, they choose to play the mixed strategy that maximises their expected reward. More formally we can write:
\begin{equation}
\hat{\sigma}^i_{BR}(\sigma^{-i})= \mathop{\rm argmax}_{\sigma^{i} \in \Delta^i} \quad r^{i}(\sigma^{i},\sigma^{-i})
\label{eq:BR}
\end{equation}    

Nash in \cite{Nash} proved that every game has at least one equilibrium which is a fixed point in the best response correspondence. A joint mixed strategy  $\hat{\sigma}$ is a Nash equilibrium iff
\begin{equation}
r^{i}(\hat{\sigma}^{i},\hat{\sigma}^{-i})\geq
r^{i}(\sigma^{i},\hat{\sigma}^{-i}) \qquad \forall  \sigma^{i} \in
\Delta^{i},i=1,\dots \mathcal{I}.
\label{eq:nash}
\end{equation}
Equation (\ref{eq:nash}) implies that no players can increase their  reward if they unilaterally change their strategy.
A special case of Nash equilibria are pure Nash equilibria. A joint action, pure strategy, where players cannot do better by
unilaterally changing their actions, is called a pure strategy Nash equilibrium.

\subsection{The TCAS problem as a symmetric game}   
The goal of a TCAS for aircraft is to avoid collision. Since all the UAVs have the same goal, it is natural to use coordination games in order to describe the task. In coordination games all players share the same reward that depends to their joint action. Thus the reward function of the game that represent a TCAS is defined as:
\begin{equation}
r(s^{i},s^{-i})=\left\{\begin{array}{rl} 
a& if s^{i}\neq s^{j} \ \forall j \in -i \\
0& otherwise
\end{array} \right.
\label{eq:reward}
\end{equation}
If we define the reward function of the UAVs using (\ref{eq:reward}), all the joint actions that are not collision free, and therefore all the cases which the UAVs fail to coordinate, are penalised and only the joint actions that solve the TCA problem produce some reward for the agents. Nevertheless there are more than one possible joint actions that lead to a positive reward, and consequently more than one pure Nash equilibria. This is due to the fact that the UAVs don't have a preference of the altitude that they want to fly at and therefore (\ref{eq:reward}) defines a symmetric game. In symmetric games someone can change the identities of the agents without changing the reward of the game, i.e in a two player game $r(s^{i},s^{-i})=r(s^{-i},s^{i})$. Thus a coordination mechanism is needed in order for the UAVs to perform their assigned task autonomously. 
\section{Coordination Mechanism}
\label{sec:coord}
\subsection{Fictitious play}
Game-theoretic learning algorithms are suitable as a coordination mechanism between the UAVs because of their small communication cost and reliability.  In fictitious play all agents choose the actions to maximise their expected reward, which is based on their estimates of their opponent's strategies, using (\ref{eq:BR}). These estimates are updated in each iteration of the game based on the observations of the other agents' actions. 

In order to compute (\ref{eq:BR}) a agent $i$ needs to know the strategies of all the other agents. But agent $i$ can only observe the actions of his opponents and not their strategies. Under the assumption that agents use a unique strategy throughout the game, opponents' strategies can be estimated using a multinomial distribution, whose parameters can be updated using the observed actions of the other agents. 

In particular, before the initial iteration of the game, each agent maintains some random non-negative weights, $\kappa_{0}$ for each of his opponents' actions. Then in each iteration of the game each agent $i$ updates his weights for the agent $j \in -i$ as follows:
\begin{equation}
\kappa_{t}^{j}(s^{j}) = \kappa_{t-1}^{j}(s^{j})+I_{s^j_t=s^j}(s^{j}),\ t=0,1,\dots, 
\label{eq:kappa}
\end{equation}
where $s^j_t$ is the observation at time $t$ and \mbox{$I_{s^j_t=s^j}(s^{j})=\left\{\begin{array}{c l}
1& if s^j_t=s^j \\
0 & otherwise.
\end{array}\right.$} The mixed
strategy of opponent $j$ is estimated then as:

\begin{equation}
\begin{array}{l l}
\sigma_{t}^{j}(s^{j}) &= \frac{\kappa_{t}^{j}(s^{j})}{\sum_{s^{j}\in S^{j}}\kappa_{t}^{j}(s^{j})} \\
&= (1-\frac{1}{t})\sigma_{t-1}^{j}(s^{j})+ \frac{1}{t}I_{s^j_t=s^j} \\
\end{array}
\label{eq:fp}
\end{equation}
Fictitious play converges to the Nash equilibrium in various categories of games, including $2\times2$ games with
generic payoffs \cite{fp2}, zero sum games \cite{fp1}, games that
can be solved using iterative dominance \cite{fp3} and potential
games \cite{fp4}. Nevertheless there are also games where fictitious play does
not converge, and becomes trapped in limit-cycles. An example of such a game is Shapley's game
\cite{shapley}. In addition in symmetric games, that we are interested in, agents may choose a mixed strategy that is a Nash equilibrium but they will actually become trapped in a cycle where their rewards will be not maximised \cite{learning_in_games}.  In the next section we introduce a new version of  estimating opponents strategies $\sigma_{t}^{j}(s^{j})$ using the Boltzmann formula. 

\subsection{Extended Kalman Filter Fictitious Play}
A variant of fictitious play, which addresses the problem of the classic algorithm in symmetric games, is the EKF-based (Extended Kalman Filter based) fictitious play \cite{ifac}.

In order to overcome difficulties that arise from the fuse of probability distributions, agents predict the unconstrained propensities \cite{cj} that their opponents have for their actions. A crucial assumption of this algorithm is that, throughout the game,  agents adapt their propensities to choose an action and based on these propensities they update their strategies and choose their actions \cite{cj}. 
\begin{figure}
\centering
\includegraphics[scale=0.25]{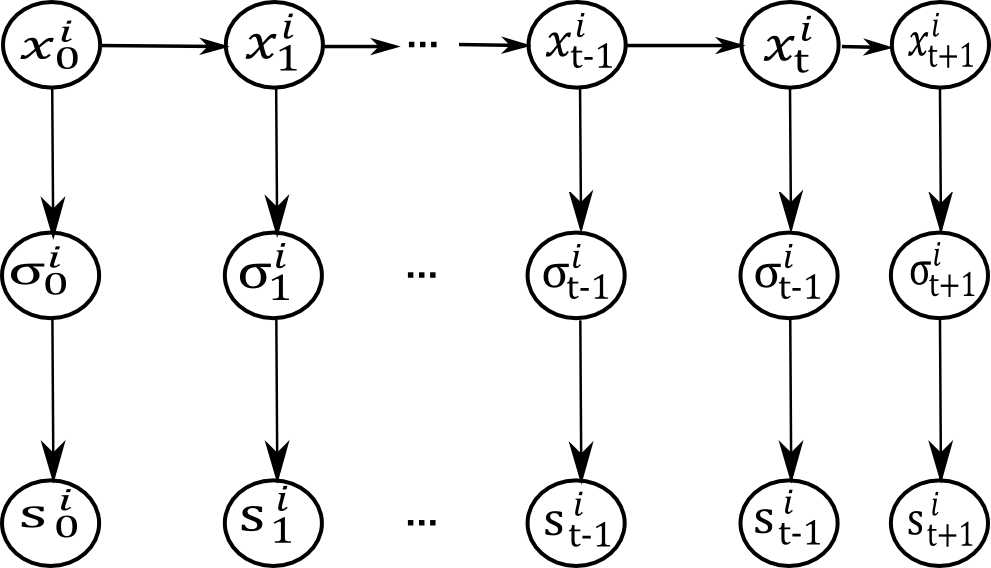}
\caption{Propensities propagation throughout the game. The Propensity at time $t$ depends only to the propensity at time $t-1$. Moreover the strategy of a agent at time $t$ depends only in the propensity of the same iteration, and the action that a agent chooses depends only at his strategy at the same iteration as well.}
\label{fig:propensities}
\end{figure}  
Figure \ref{fig:propensities} depicts how propensities of agent $i$ change through the iterations of the game and how they are related to strategies and actions.
Based on the fact that agents have no prior knowledge about their opponents' strategies, an autoregressive model is chosen in order to propagate the propensities \cite{cj}. In addition, inspired from the sigmoid functions that are used in neural networks to connect the weights and the observations, a Boltzman formula (\ref{boltz}) is used to relate the propensities with opponents' strategies \cite{cbishop}. Thus the following state space model is used to describe the fictitious play process:
\begin{align}
\label{eq:fpekf}
x^j_{t}&=x^j_{t-1}+\xi^j_{t-1} \nonumber  \\ 
I_{s^j_t=s^j}(s^{j})&= h(x^j_{t})+\zeta^j_{t},\ j\in \{1,...,\mathcal{I}\}\backslash \{i\} 
\end{align}
where the components of $h$ are 
\begin{equation} h_k(x)=\frac{exp(x_k/\tau)}{\sum_k exp(x_k/\tau)},\ \ k\in  S^j \label{boltz}\end{equation} 
$\xi^j_{t-1} \sim N(0,\Xi)$, is the noise of the propensity  process which comprises the internal states and $\zeta_{t} \sim N(0,Z)$  is the error of the observation of propensities by the indicator function with zero mean and covariance matrix $Z$, which occurs because we observe a discrete 0-1 process, such as the best response in (\ref{eq:BR}) through the  continuous Boltzmann formula $h(\cdot)$  in (\ref{boltz}) in which $\tau$ is a "temperature parameter". 

agent $i$ then evaluates his opponents strategies using his estimates as:
\begin{equation}
 \sigma^j_{tk}=\frac{exp(\bar{x}^j_{kt} / \tau)}{\sum_k exp(\bar{x}^j_{kt} / \tau)}.
\label{eq:strategies}
\end{equation}
where $\bar{x}^j_{kt}$ is agent $i$'s prediction of the propensities of opponent $j$ to choose action  $k$ based  the state equations in (\ref{eq:fpekf}) and using observations up to time $t-1$. Agent $i$ then uses the estimates of its opponents strategies in (\ref{eq:strategies}) to choose an action by best response (\ref{eq:BR}) evaluation. The EKF estimation is done by any standard textbook procedure \cite{bayes_filter}. 


Table \ref{skata} summarises the fictitious play algorithm when EKF is used to predict opponents strategies. 

\begin{table*}
\begin{center}
\begin{tabular}{p{12cm}}
\hline
\hline
\begin{enumerate}
\item At time $t$ agent $i$ maintains   estimates  of its opponent's propensities up to time $t-1$, $\hat{x}^j_{t-1}\ ,$, with covariance $P^j_{t-1}$ of its distribution.

\item Agent $i$ predicts its estimates about its opponents' propensities to $\bar{x}_tk^j,\ j\in \{1,...,\mathcal{I}\}\backslash\{i\},\ k\in S^j$  using the state equations in (\ref{eq:fpekf}).

\item Based on the propensities in 2 each agent updates its beliefs about its
opponents' strategies using $\sigma_{tk}^{j}=\frac{exp(\bar{x}^j_{tk}/\tau)}{\sum_{k}exp(\bar{x}^j_{tk}/\tau)},\ k\in S^j$.

\item Agent $i$ chooses an action based on the beliefs in 3 and applies best response decision rule.

\item The agent $i$  observes its opponents' actions $s^j_{tk},\ j\in \{1,...,\mathcal{I}\}\backslash\{i\}$.

\item The agent update its estimates of all of its opponents' propensities using extended Kalman Filtering to obtain $\hat{x}^j_t,\ j\in \{1,...,\mathcal{I}\}\backslash\{i\}$.
\end{enumerate}
\\
\hline
\hline
\end{tabular}
\caption{EKF based fictitious play algorithm.}
\label{skata}
\end{center}
\end{table*}

\section{Experimental evaluation}
\label{sec:exper}

The scenario presented in this paper requires autonomous control of a pair of quadcopters, in this case the model selected is the Parrot AR.Drone 2.0. The quadcopter has one forward-facing and one downward-facing camera. This camera will be used for location of the parter quadcopter, and for navigation to maintain position within the room. 

\subsection{Quadcopter Driver}
The autonomy package\footnote{\url{https://github.com/AutonomyLab/ardrone autonomy
}} for the AR.Drone developed for the Robot Operating System~\cite{Quigley} provides the main harness for communicating with the quadcopter, and provides an interface to the quadcopter navigation data.
 
\subsection{Maintaining Quadcopter Position} \label{sec:slam}
\begin{figure}
\centering
\subfigure[Sample Room.] {
\includegraphics[width=.4\textwidth,keepaspectratio=true]{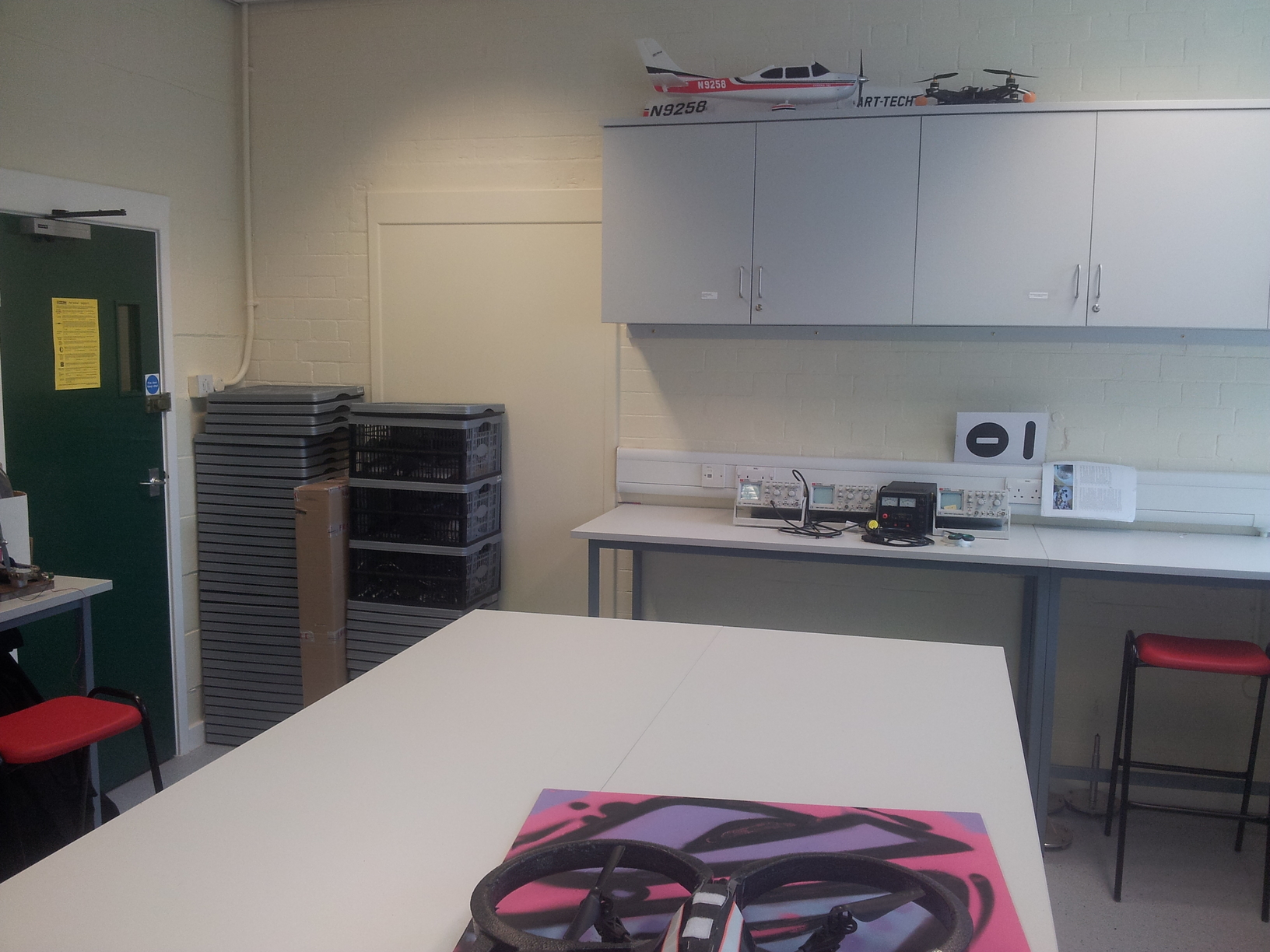}
\label{Figure:room}
}
\subfigure[3D View of Keypoints.] {
\includegraphics[width=.5\textwidth,keepaspectratio=true]{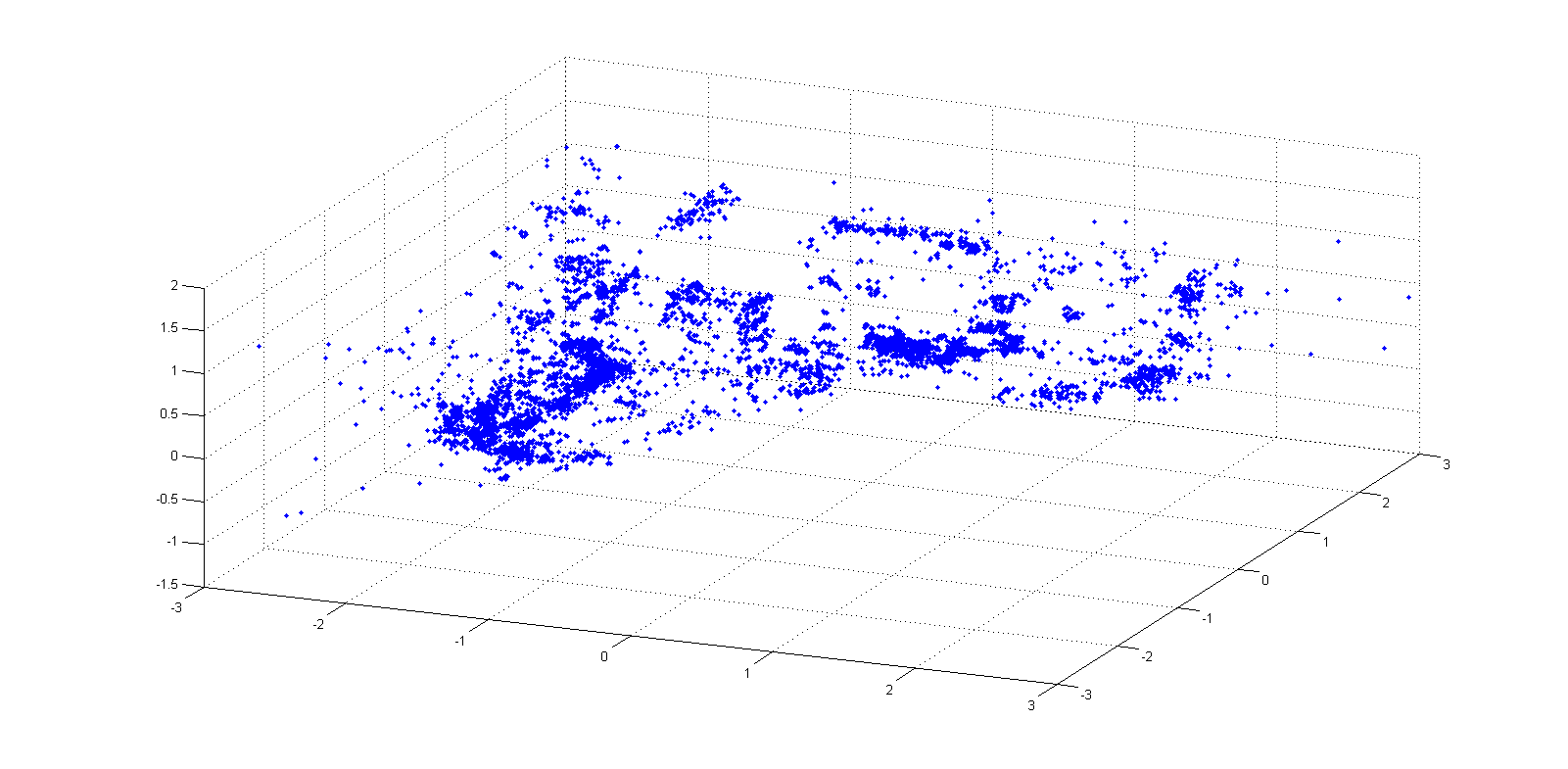}
\label{Figure:xyz_scatter}
}
\subfigure[2D View of Keypoints from Top Down.] {
\includegraphics[width=.5\textwidth,keepaspectratio=true]{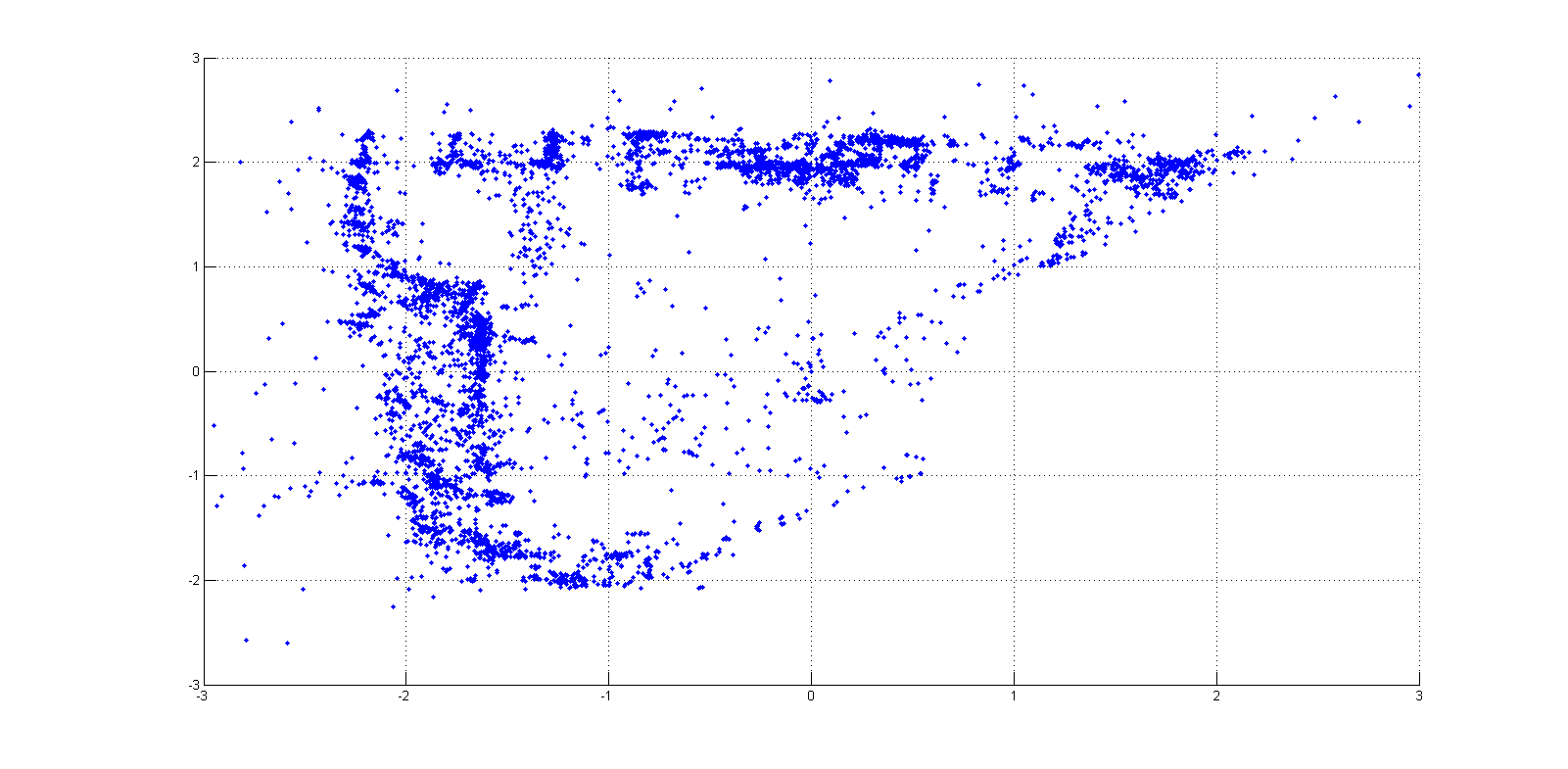}
\label{Figure:xy_scatter}
}
\caption{Keypoints Identified within a Room Showing Distribution Across a Scene.}
\label{Figure:keypoints}
\end{figure}

In order to maintain position this paper utilises the Technical University of Munich (TUM) AR.Drone package~\cite{Engel12,Engel14}. This algorithm provides a monocular Simultaneous Location and Mapping (SLAM) based on the Parallel Tracking and Mapping (PTAM) algorithm~\cite{Klein}. This algorithm uses a collection of keypoints identified within a stream of images to track the position of a camera relative to a scene. Typically these keypoints are object edges or corners which helps build a picture of the environment as shown in Figure~\ref{Figure:keypoints}.

By coupling the tracking of these keypoints to information coming from the onboard sensors, The TUM AR.Drone package can produce a 3D map of the world locating these keypoints in space by tracking them between frames, using a pair of initialisation frames using monocular SLAM. Once these keypoints are known, the quadcopter position can be fixed in space, relative to the keypoints. This provides a very robust method for positioning the quadcopter. Incorporated within the package is an autopilot function that takes these 3D coordinates as demand positions for the quadcopter.

By incorporating the autopilot and the SLAM routine each quadcopter can be initialised into position to begin the experiment. Appropriate commands can then be passed to the autopilot as a result of the decision making process.

\subsection{Quadcopter Recognition}
Quadcopter recognition is performed using inbuilt functionality of the Parrot AR.Drone 2.0. The front facing cameras processes images and has been configured to locate a colour pattern, blue-orange-blue. As part of the navigation data, the quadcopter reports the presence of the colour pattern in the current field of view. Each quadcopter has the colour pattern affixed to their front-facing side, adjacent to the forward-facing camera. Therefore since each quadcopter knows its own altitude, by observing or not observing the other quadcopter it can infer its action.

\subsection{Mission Management}
\begin{figure}[ht!]
\centering
\includegraphics[width=.75\textwidth,keepaspectratio=true]{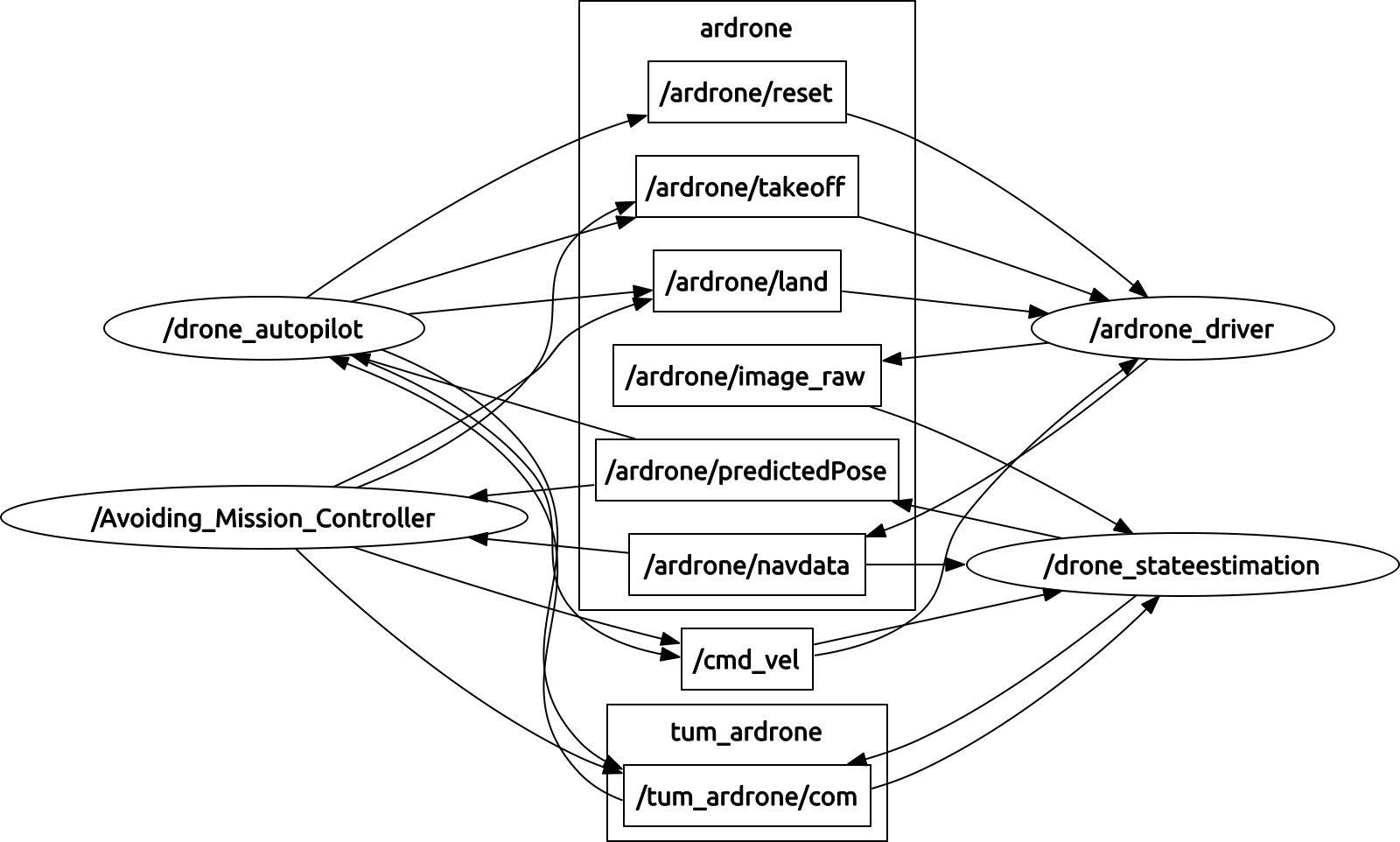}
\caption{Underlying ROS Connection Graph.}
\label{Figure:ros_graph}
\end{figure}

A mission management node contains the intelligence for the mission. Figure~\ref{Figure:ros_graph} shows the connection into the complete ROS graph for the system. The mission management node takes decisions based on the observed environment, therefore interfaces to the navigation data to receive tag information, and information on the predicted pose to understand attitude and orientation provided by the SLAM system outlined in Section~\ref{sec:slam}. At the beginning of each experiment both quadcopters have their SLAM systems initialised separately to ensure that they do not capture the other as part of the background. The mission node controls takeoff and landing directly through connected topics within the driver, but uses the TUM AR.Drone command channel to communicate with the autopilot to adjust quadcopter position.

The mission management node controls the safe execution of the manoeuvre. The quadcopter starts by observing whether the other player is visible. Every 8s a decision is taken based on the algorithm outlined in Section~\ref{sec:decision_making}, the result of this decision is passed to the autopilot to change the quadcopter position appropriately; either rising 1m if in the low position, descending 1m if in the upper position or holding station. During this 8s window, if the quadcopter observers the absence of the other player for a continuous 4s period it begins the passing procedure by travelling forwards, adjusting the longitudinal position supplied to the autopilot moving it forwards whilst maintaining the current altitude.

\subsection{A symmetric Game and the Decision Making Process} \label{sec:decision_making}
In our implementation we allowed the quadcopters to fly in opposite directions using two altitudes, high and low altitude hereafter. When (\ref{eq:reward}) is used, players share the rewards of the game which is presented in Table \ref{tab:rewards}. 

\begin{table}
\centering
\begin{tabular}{|c|c|c|}
\hline
&High altitude&Low altitude \\ \hline
Low altitude& 1&0 \\ \hline
High altitude&0&1 \\ \hline 
\end{tabular}
\caption{The reward that the two quadcopters share. }
\label{tab:rewards}
\end{table}
We arbitrarily chose $a=1$, but the presented results will be valid for any other value of constant $a$.

The rest of this section provides details about the implementation of EKF fictitious play in the proposed TCAS framework. We consider the case where the above mentioned symmetric game has been played for $t-1$ iterations and the players use EKf fictitious play to update their estimations of their opponent's strategy. 

We will present the updates of only one robot since the updates of the other quadcopter will be identical. In our implementation we used the following parameters for $\Xi$ and $Z$ and $\tau$: \newline
$\Xi=\left[\begin{array}{cc}
0.05& 0\\
0& 0.05\\
\end{array} \right], \quad Z=\left[\begin{array}{cc}
0.3& 0\\
0& 0.3\\
\end{array} \right], \quad \tau=2$ 

From the previous iterations of the game the UAV will have an updated estimation of $p(x_{t-1}|s_{1:t-1})$ in the light of joint action $s_{t-1}$. Therefore since it will approximate $p(x_{t-1}|a_{1:t-1})$, using $\hat{x}_{t-1}$, the variance of this approximation will be  $P_{t-1}$. For the rest of this paper the $i_{th}$ element of a vector $y$ will be denoted as $y[i]$ and the $(i,j)_{th}$ element of a matrix $y$ as $y[i][j]$. Without loss of generality we assume that  $\hat{x}[1]$ is the approximation of the propensity the other UAV has to fly in high altitude and $\hat{x}[2]$ is the approximation of the propensity the other UAV has to fly in low altitude. Consequently $P[1][1]$ and $P[2][2]$ is the variance of the approximation of the propensity the other UAV has to fly in high and low altitude  altitude respectively. The covariance between the two actions propensities is estimated in $P[1][2]=P[2][1]$. 

Then at time $t$ before it makes a decision it needs to approximate $p(x_{t}|s_{1:t-1})$. The approximation of the propensities and its variance is: 
\begin{displaymath}
\begin{array}{rl}
\hat{x}^{-}_{t}[1]=&\hat{x}_{t-1}[1] \\
\hat{x}^{-}_{t}[2]=&\hat{x}_{t-1}[2]
\end{array}
\end{displaymath}

\begin{displaymath}
\begin{array}{r l}
P^{-}_{t}[1][1]=&P_{t-1}[1][1]+\Xi[1][1]+d\\
P^{-}_{t}[1][2]=&P_{t-1}[1][2]\\
P^{-}_{t}[2][1]=&P_{t-1}[2][1]\\
P^{-}_{t}[2][2]=&P_{t-1}[2][2]+\Xi[2][2]+d\\
\end{array}
\end{displaymath}
where $d=0.1+0.0001\lvert\mathbf{n}\rvert$, $\mathbf{n} \sim N(0,0.0001)$
The estimation of of the other UAV's strategy is 

\begin{displaymath}
\begin{array}{c c}
\sigma_{t}[1]&=\frac{\exp(\hat{x}^{-}_{t}[1]/\tau)}{\exp(\hat{x}^{-}_{t}[1]/\tau)+\exp(\hat{x}^{-}_{t}[2]/\tau)}\\
\sigma_{t}[2]&=\frac{\exp(\hat{x}^{-}_{t}[2]/\tau)}{\exp(\hat{x}^{-}_{t}[1]/\tau)+\exp(\hat{x}^{-}_{t}[2]/\tau)}\\
\end{array}
\end{displaymath}

Then based of this estimation the UAV make a decision about the altitude that it will choose. Because of the symmetry of the game and the fact that Best responce (\ref{eq:BR}), the UAV chooses the action which has smaller to be selected by the other UAV, and thus $s_{t}^{i}=argmin_{j=1,2} \sigma_{t}[j]$.

After both UAVs choosing their actions then they should update their estimations by estimating $p(x_{t}|s_{1:t})$. Without loss of generality we will assume that the UAV which we estimate its strategy chose to fly in high altitude, action 1. The approximation of $p(x_{t}|s_{1:t})$ and its covariance, based on the update step of the EKF process are as follows: 

The difference between the estimation and the observed action is evaluated as: 
\begin{displaymath}
\begin{array}{rl}
v_{t}[1]=&\sigma_{t}[2]\\
v_{t}[2]=&-\sigma_{t}[2]\\
\end{array}
\end{displaymath}

The Jacobian of the transformation function $h$ is computed as: 
\begin{displaymath}
H_{t}=\left[
\begin{array}{c c}
\sigma_{t}[1]\sigma_{t}[2] & -\sigma_{t}[1]\sigma_{t}[2] \\
-\sigma_{t}[1]\sigma_{t}[2]&\sigma_{t}[1]\sigma_{t}[2]\\
\end{array}\right]
\end{displaymath}
The variance of the approximation process is 
\begin{displaymath}
S_{t}=c_{1}\left[
\begin{array}{c c}
1 & -1 \\
-1&1\\
\end{array}\right]+\left[\begin{array}{c c}
(1/t)& 0 \\
0&(1/t)\\
\end{array}\right]
\end{displaymath}
where $c_{1}=\sigma_{t}[1]\sigma_{t}[2](P_{t}^{-}[1][1]+P_{t}^{-}[2][2])-2P_{t}^{-}[1][2])$. The Kalman gain is computed as: 
\begin{displaymath}
K_{t}=c_{2}\left[
\begin{array}{c c}
P[1][1]-P[1][2] & P[1][2]-P[1][1] \\
P[1][2]-P[2][2]&P[2][2]-P[1][2]\\
\end{array}\right]
\end{displaymath}
where $c_2=\frac{1}{tc_1}\frac{1}{P_{t}^{-}[1][1]+P_{t}^{-}[2][2])-2P_{t}^{-}[1][2]}\frac{1}{c_{1}(1+c_{1})^{2}}$. 
Finally the  updates for $\hat{x}_{t}$ and $P_{t}$ are:
\begin{displaymath}
\begin{array}{rl}
\hat{x}_{t}[1]=&\hat{x}_{t}^{-}[1]+2(P[1][1]-P[1][2])\sigma_{t}[2]c_{2} \\
\hat{x}_{t}[2]=&\hat{x}_{t}^{-}[2]-2(P[1][1]-P[1][2])\sigma_{t}[2]c_{2}
\end{array}
\end{displaymath}

\begin{displaymath}
\begin{array}{r l}
P_{t}[1][1]=&P_{t-1}^{-}[1][1]-(2+\frac{1}{t})c_{2}^{2}c_{1}(c_{3}^{2}-c_{3}c_{4})\\
P_{t}[1][2]=&P_{t-1}^{-}[1][2]-(2+\frac{1}{t})c_{2}^{2}c_{1}(-c_{3}^{2}-c_{3}c_{4})\\
P_{t}[2][1]=&P_{t-1}^{-}[2][1]-(2+\frac{1}{t})c_{2}^{2}c_{1}(-c_{3}^{2}-c_{3}c_{4})\\
P_{t}[2][2]=&P_{t-1}^{-}[2][2]-(2+\frac{1}{t})c_{2}^{2}c_{1}(c_{4}^{2}-c_{3}c_{4})\\
\end{array}
\end{displaymath}
where $c_{3}=P[1][1]-P[1][2]$ and $c_{4}=P[2][2]-P[1][2]$.

\begin{figure}
\includegraphics[width=\columnwidth]{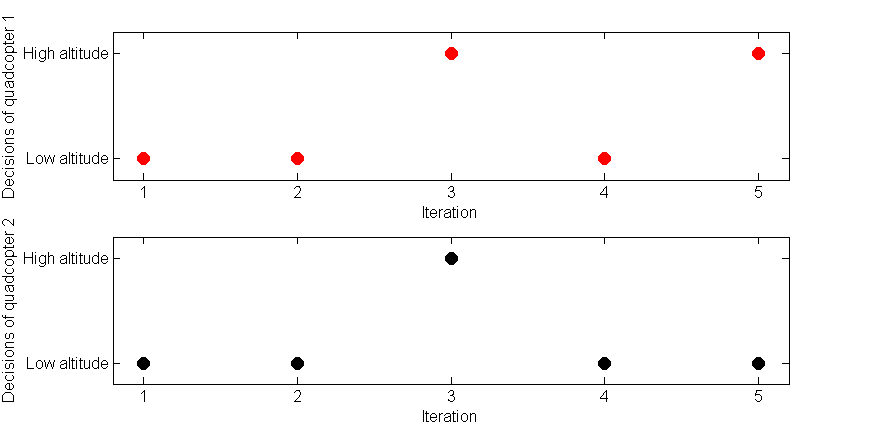}
\caption{Decisions of the quadcopters in the TCAS implementation.}
\label{fig:actions}
\end{figure}

\begin{figure}
\centering
\subfigure[UAVs are changing simultaneously their altitude to low altitude.] {
\includegraphics[width=.475\textwidth,keepaspectratio=true]{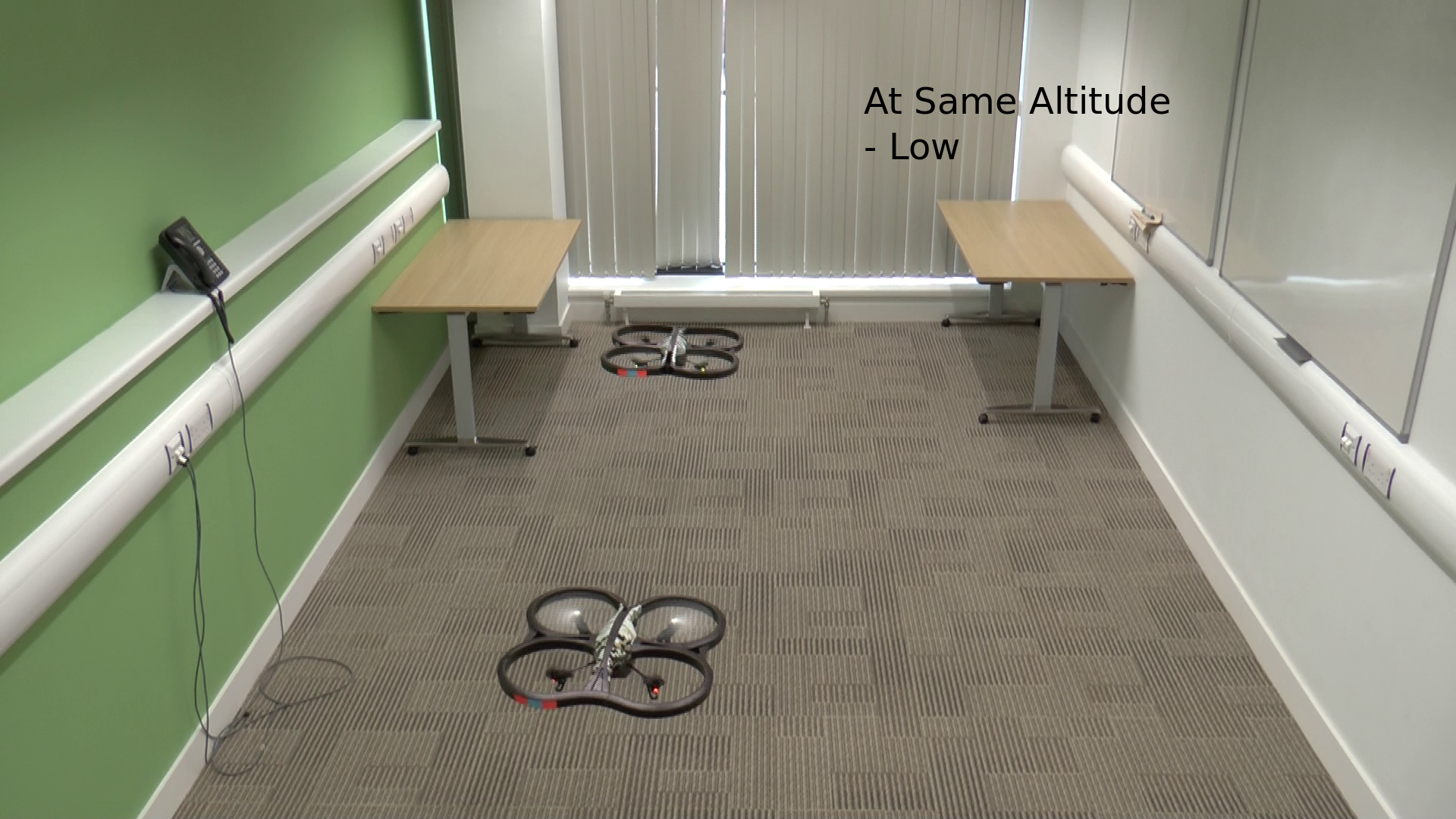}
\label{Figure:samealt_low}
}
\subfigure[UAVs are changing simultaneously their altitude to high altitude.] {
\includegraphics[width=.475\textwidth,keepaspectratio=true]{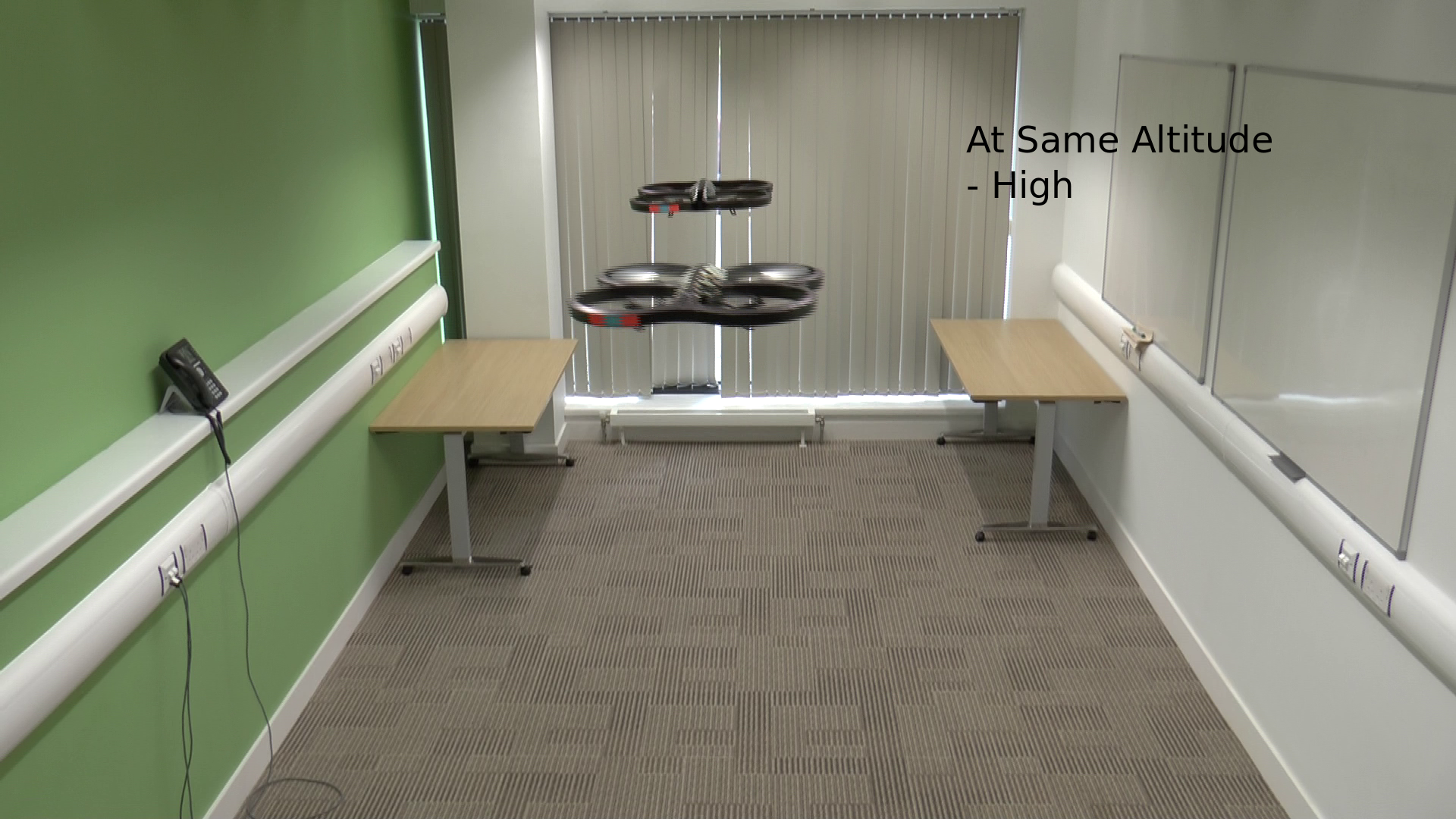}
\label{Figure:samealt_high}
}
\subfigure[UAVs manage to coordinate, one fly in high and the other in low altitude.] {
\includegraphics[width=.475\textwidth,keepaspectratio=true]{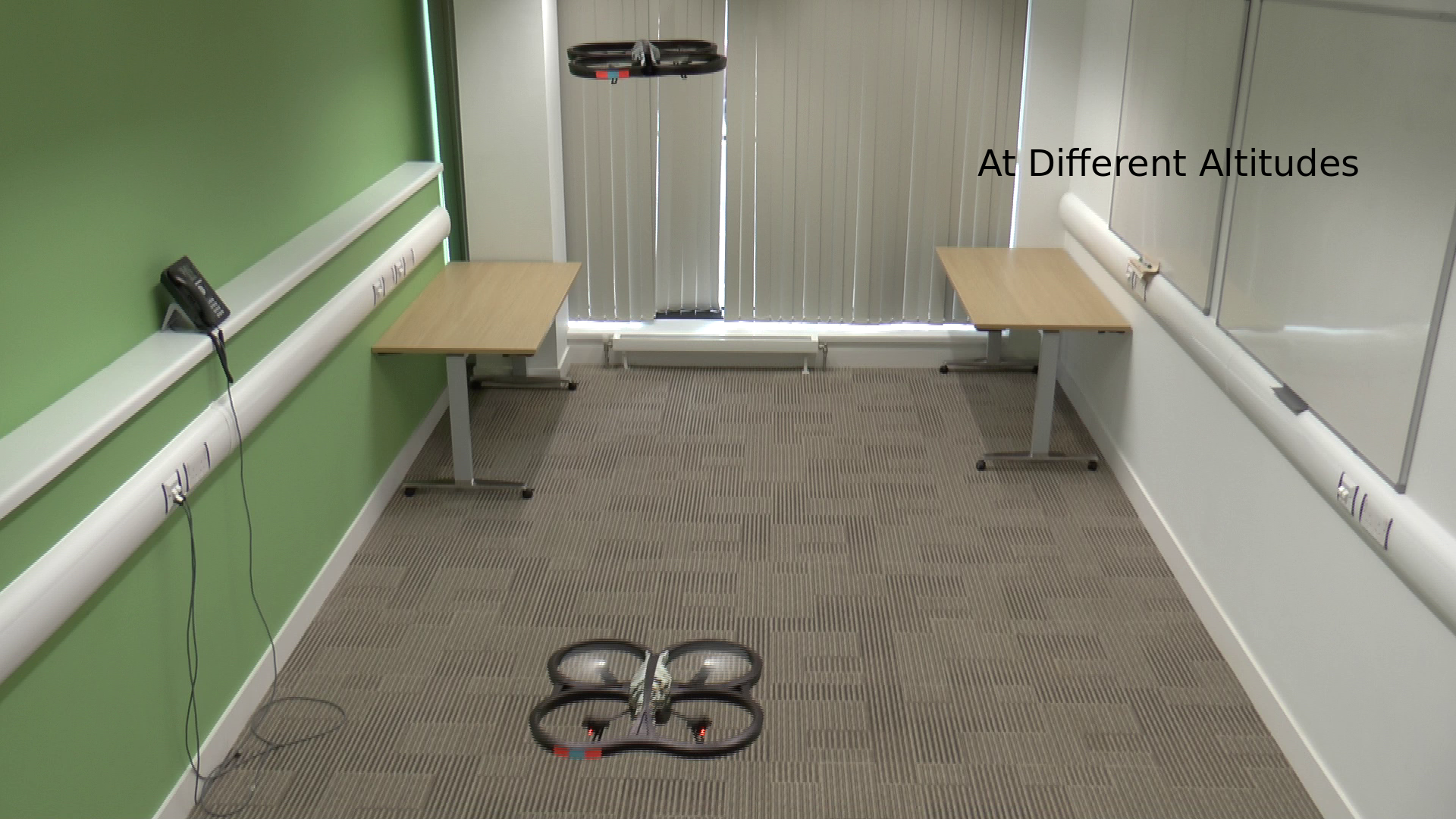}
\label{Figure:different_alt}
}
\subfigure[UAVs are flying in different directions, successfully passing.] {
\includegraphics[width=.475\textwidth,keepaspectratio=true]{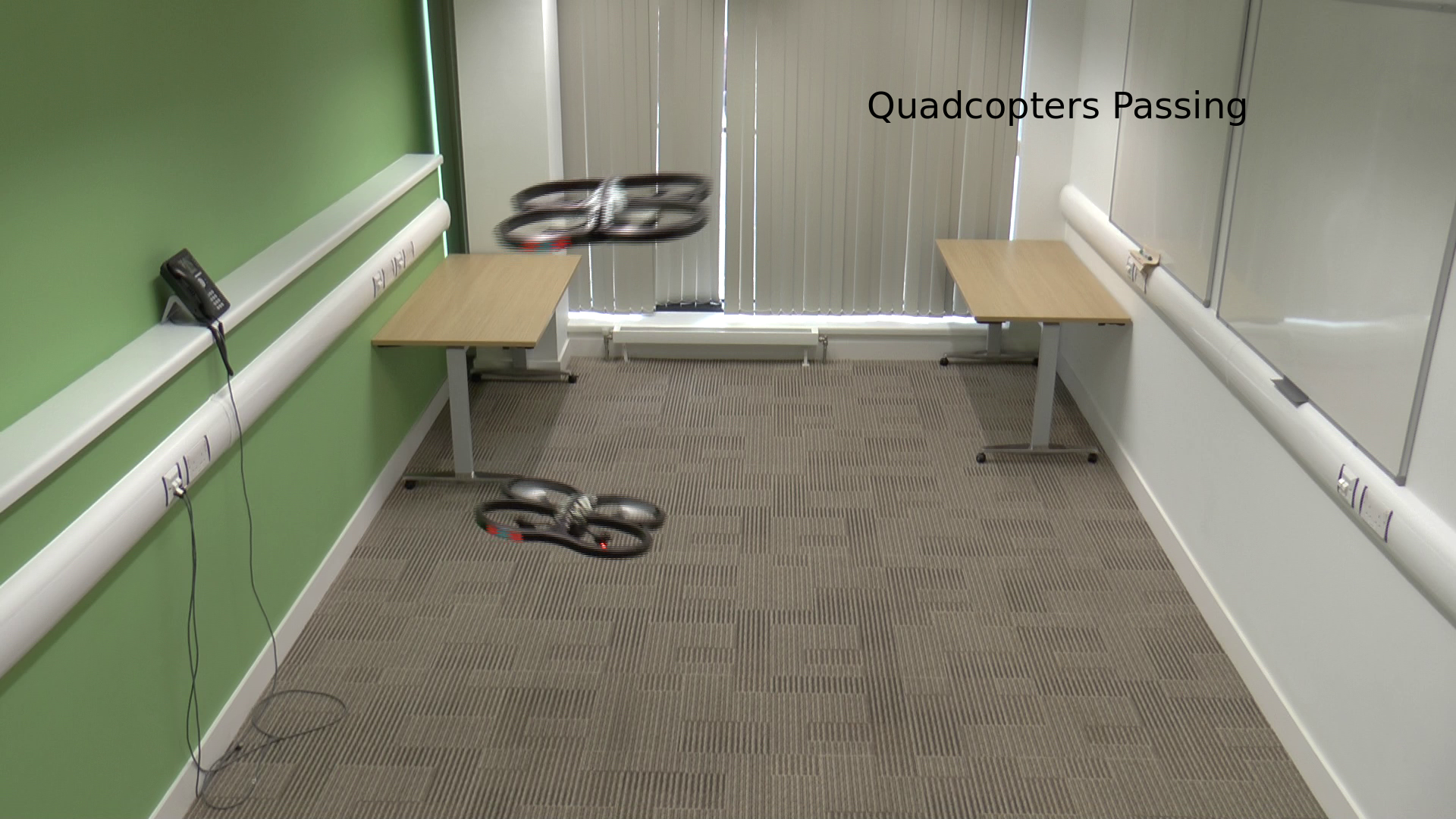}
\label{Figure:qc_passing}
}
\caption{Snapshots of the Implementation of Two Quadcopters Safely Passing Avoiding a Collision.}
\label{Figure:video_results}
\end{figure}

Based on the above described process the UAVs made the decisions that are depicted in Figure \ref{fig:actions}. Initially both UAVs choose to flight in low altitude. They need 2 iterations to learn the other UAV's action and then they change their action which is the best response to low altitude. Nevertheless they both change altitude and thus the should keep play the symmetric game. In the next iteration of the EKF fictitious play they change together their actions to low altitude. This time quadcopter 2 adapt its estimation of quadcopter 1's propensity slower than quadcopter 1 does and only  quadcopter 1 change altitude. In the new altitudes they don't observe the other UAV and they choose to complete their task to fly in opposite directions. Figures~\ref{Figure:samealt_low},~\ref{Figure:samealt_high},~\ref{Figure:different_alt} and \ref{Figure:qc_passing} depict snapshots of the implementation of the above algorithm using two Parrot AR.Drone 2.0.

\section{Conclusions}
\label{sec:concl}
A new TCAS principle has been introduced for unmanned aerial vehicles represented by agents capable of decisions about the flight of their planes, based on a variant of fictitious-play-based symmetric game with no communication requirements between the agents. The agents can infer all the necessary information they need by observing the other agent's actions. They use  a variant of fictitious play, which combines the classical algorithm with extended Kalman filters, as coordination mechanism in order  to avoid collision and accomplish their task.



\end{document}